\documentclass[preprint,times,twocolumn,final,authoryear]{elsarticle}

\usepackage{framed,multirow}

\usepackage{amssymb, amsmath}
\usepackage{latexsym}

\usepackage{url}
\usepackage{xcolor}
\definecolor{newcolor}{rgb}{.8,.349,.1}

\usepackage{epsfig}

\usepackage{float}

\journal{Computer Vision and Image Understanding}

\newcommand\blfootnote[1]{%
  \begingroup
  \renewcommand\thefootnote{}\footnote{#1}%
  \addtocounter{footnote}{-1}%
  \endgroup
}

\begin{document}

\clearpage

\ifpreprint
  \setcounter{page}{1}
\else
  \setcounter{page}{1}
\fi

\begin{frontmatter}

\title{Multitask Learning for Large-scale Semantic Change Detection}

\author[1,2]{Rodrigo {Caye Daudt}} 
\author[1]{Bertrand {Le Saux}}
\author[1]{Alexandre {Boulch}}
\author[2]{Yann {Gousseau}}

\address[1]{DTIS, ONERA, Universit\'{e} Paris-Saclay, FR-91123 Palaiseau, France}
\address[2]{LTCI, T\'{e}l\'{e}com ParisTech, FR-75013 Paris, France}

\begin{abstract}
Change detection is one of the main problems in remote sensing, and is essential to the accurate processing and understanding of the large scale Earth observation data available. Most of the recently proposed change detection methods bring deep learning to this context, but change detection labelled datasets which are openly available are still very scarce, which limits the methods that can be proposed and tested. In this paper we present the first large scale very high resolution semantic change detection dataset, which enables the usage of deep supervised learning methods for semantic change detection with very high resolution images. The dataset contains coregistered RGB image pairs, pixel-wise change information and land cover information. We then propose several supervised learning methods using fully convolutional neural networks to perform semantic change detection. Most notably, we present a network architecture that performs change detection and land cover mapping simultaneously, while using the predicted land cover information to help to predict changes. We also describe a sequential training scheme that allows this network to be trained without setting a hyperparameter that balances different loss functions and achieves the best overall results.
\end{abstract}

\end{frontmatter}


\section{Introduction\protect\blfootnote{This work was originally submitted under the title "High Resolution Semantic Change Detection", and had its title changed during the review process.}}

One of the main purposes of remote sensing is the observation of the evolution of the land. Satellite and aerial imaging enables us to keep track of the changes that occur around the globe, both in densely populated areas as well as in remote areas that are hard to reach. That is why change detection is a problem so closely studied in the context of remote sensing~\citep{coppin2004review}. Change detection is the name given to the task of identifying areas of the Earth's surface that have experienced changes by jointly analysing two or more coregistered images~\citep{bruzzone2013novel}. Changes can be of several different types depending on the desired application, e.g. those caused by natural disasters, urban expansion, and deforestation. In this paper we treat change detection as a dense classification problem, aiming to predict a label for each pixel in an input image pair, i.e. achieving semantic segmentation.

The search for ever more accurate change detection comes from the value of surveying large amounts of land and analysing its evolution over a period of time. Detecting changes manually is a slow and laborious process~\citep{singh1989review} and the problem of automatic change detection using image pairs or sequences has been studied for many decades. The history of change detection algorithms and overviews of the most important methods are described in the reviews \citet{singh1989review} and \citet{hussain2013change}. Throughout the years, change detection benefited a lot from computer vision and image processing advances. In recent years, computer vision made tremendous progress thanks to machine learning techniques, and these were used for solving a wide range of problems related to image understanding~\citep{lecun2015deep}.

The rise of these techniques is explained by three main factors. First, the hardware required for the large amounts of calculations that are often required for machine learning techniques is becoming cheaper and more powerful. Second, new methods are being proposed to exploit the data in innovative ways. Finally, the amount of available data is increasing, which is essential for many machine learning techniques.

In this paper we propose a versatile supervised learning method to perform pixel-level change detection from image pairs based on state-of-the-art computer vision ideas. The proposed method is able to perform both binary and semantic change detection using very high resolution (VHR) images. Binary change detection attempts to identify which pixels correspond to areas where changes have occurred, whereas semantic change detection attempts to further identify the type of change that has occurred at each location. The proposed method is able to perform change detection using VHR images from sources such as WorldView-3, Pl\`{e}iades and IGN's BD ORTHO. As was described by \citet{hussain2013change} and \citet{bruzzone2013novel}, VHR change detection involves several extra challenges.

A new VHR semantic change detection dataset of unprecedented size is also presented in this paper. This dataset will be released publicly to serve as a benchmark and as a research tool for researchers working on change detection. The methods used to create this dataset, as well as the limitations of the available data, will be described later on. Until now, the most advanced ideas brought to computer vision by deep learning techniques could not be applied to change detection due to the lack of large annotated datasets. This dataset will enable the application of more sophisticated machine learning techniques that were heretofore too complex for the amount of change detection data available.

\section{Related work}

The work presented in this paper is based on several different ideas coming from two main research areas: change detection and machine learning. This section contains a discussion about the works hat have more heavily influenced this work, providing details about unsupervised methods, supervised learning, and fully convolutional networks for semantic segmentation.

\textbf{Change detection} algorithms usually comprise two main steps~\citep{singh1989review, hussain2013change}. First, a difference metric is proposed so that a quantitative measurement of the difference between corresponding pixels can be calculated. The image generated from this step is usually called a difference image. Second, a thresholding method or decision function is proposed to separate the pixels into "change" and "no change" based on the difference image. These two steps are usually independent. Post-processing and pre-processing methods are sometimes used to improve results. Many algorithms use out-of-the-box registration algorithms and focus on the other main steps for change detection~\citep{hussain2013change}. Most papers on change detection propose either a novel image differencing method \citep{bovolo2005wavelet, el2016convolutional, el2017zoom, zhan2017change} or a novel decision function \citep{bruzzone2000automatic, celik2009unsupervised}. A well established family of change detection methods is change vector analysis (CVA), considering the multispectral difference vector in polar or hyperspherical coordinates and attempting to characterise the changes based on the associated vectors at each pixel~\citep{lambin1994change, bovolo2007theoretical, hussain2013change}. Most methods that propose image differencing techniques followed by thresholding assume that a threshold is chosen based on the difference image. The authors of \citet{hussain2013change} and \citet{rosin2003evaluation} noted that the performance of such algorithms is scene dependent.

\begin{figure*}[ht]
        \centering
        \begin{minipage}[b]{0.13\linewidth}
            \centering
            \centerline{\epsfig{figure=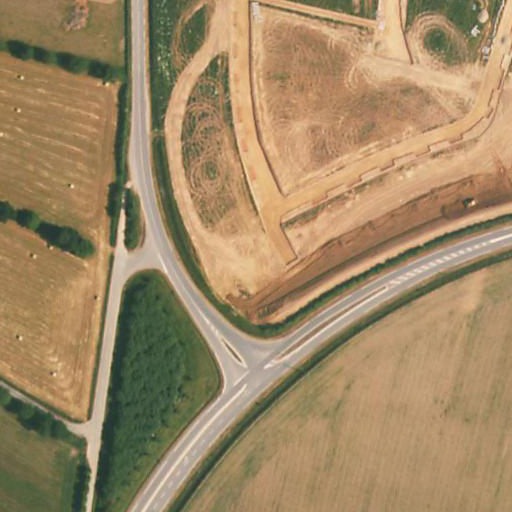,width=\linewidth}}
            \centerline{(a) Image 1.}\medskip
        \end{minipage}
        \hfill
        \begin{minipage}[b]{0.13\linewidth}
            \centering
            \centerline{\epsfig{figure=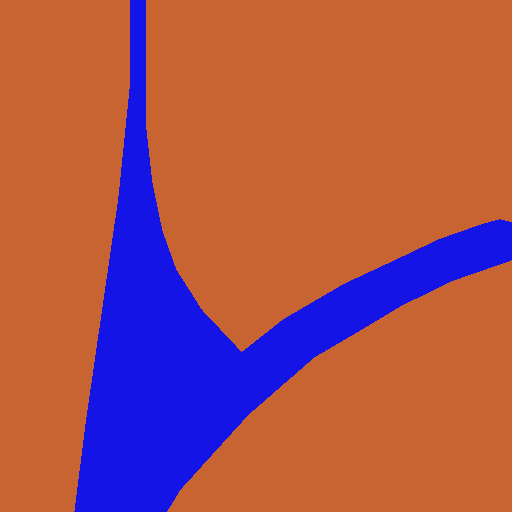,width=\linewidth}}
            \centerline{(b) LCM 1.}\medskip
        \end{minipage}
        \hfill
        \begin{minipage}[b]{0.13\linewidth}
            \centering
            \centerline{\epsfig{figure=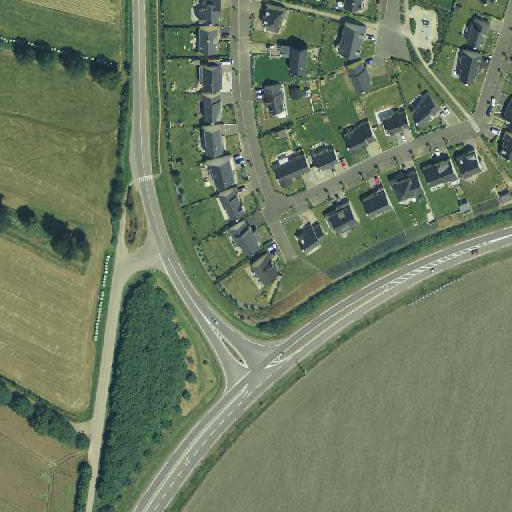,width=\linewidth}}
            \centerline{(c) Image 2.}\medskip
        \end{minipage}
        \hfill
        \begin{minipage}[b]{0.13\linewidth}
            \centering
            \centerline{\epsfig{figure=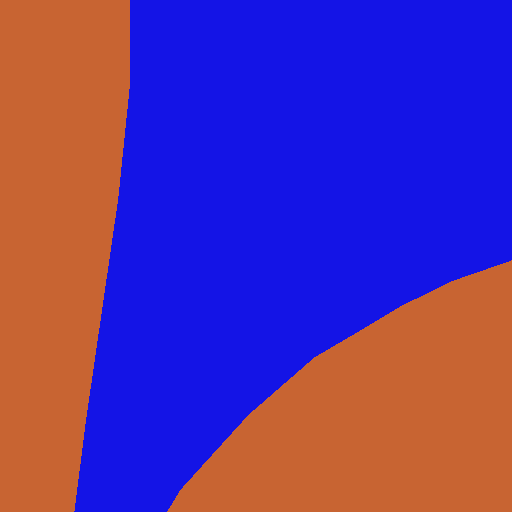,width=\linewidth}}
            \centerline{(d) LCM 2.}\medskip
        \end{minipage}
        \hfill
        \begin{minipage}[b]{0.13\linewidth}
            \centering
            \centerline{\epsfig{figure=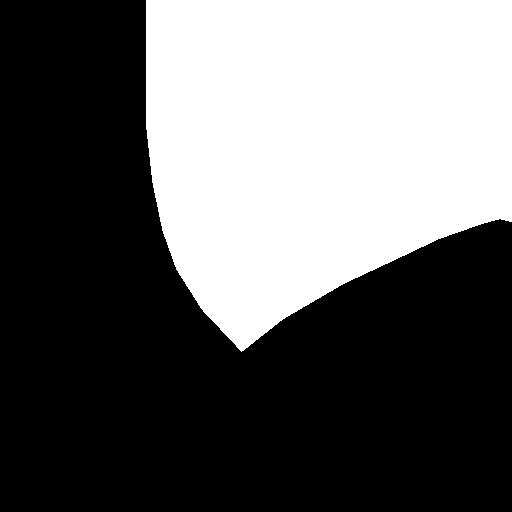,width=\linewidth}}
            \centerline{(e) Change map.}\medskip
        \end{minipage}
        \\
        \begin{minipage}[b]{0.13\linewidth}
            \centering
            \centerline{\epsfig{figure=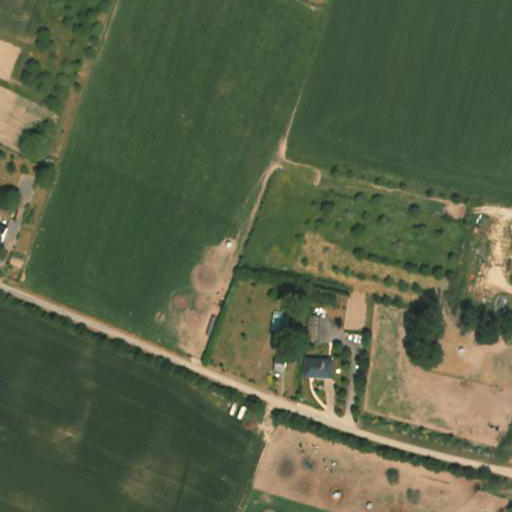,width=\linewidth}}
            \centerline{(f) Image 1.}\medskip
        \end{minipage}
        \hfill
        \begin{minipage}[b]{0.13\linewidth}
            \centering
            \centerline{\epsfig{figure=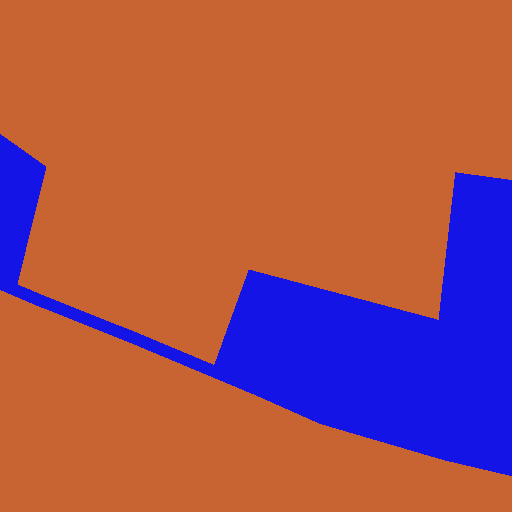,width=\linewidth}}
            \centerline{(g) LCM 1.}\medskip
        \end{minipage}
        \hfill
        \begin{minipage}[b]{0.13\linewidth}
            \centering
            \centerline{\epsfig{figure=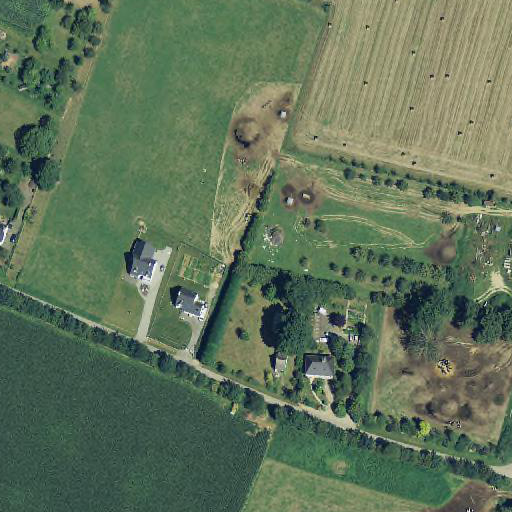,width=\linewidth}}
            \centerline{(h) Image 2.}\medskip
        \end{minipage}
        \hfill
        \begin{minipage}[b]{0.13\linewidth}
            \centering
            \centerline{\epsfig{figure=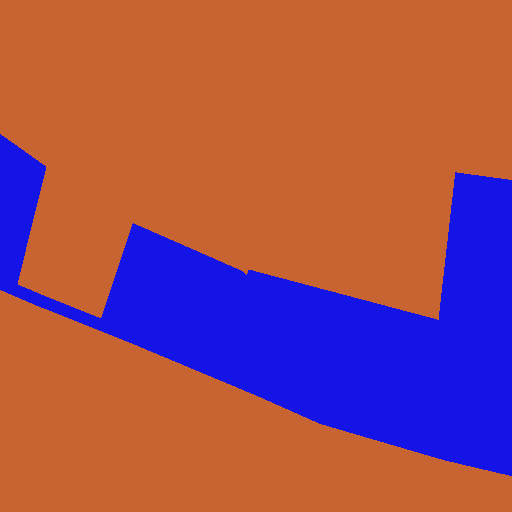,width=\linewidth}}
            \centerline{(i) LCM 2.}\medskip
        \end{minipage}
        \hfill
        \begin{minipage}[b]{0.13\linewidth}
            \centering
            \centerline{\epsfig{figure=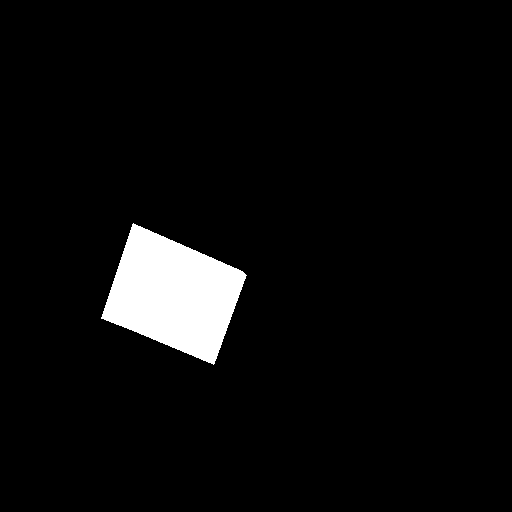,width=\linewidth}}
            \centerline{(j) Change map.}\medskip
        \end{minipage}
        \caption{\label{fig:example}Examples of image pairs, land cover maps (LCM) and associated pixel-wise change maps from the HRSCD dataset. In the depicted LCMs, blue represents the "artificial surfaces" class, and orange represents the "agricultural areas" class.}
\end{figure*}

\citet{hussain2013change} categorise change detection algorithms into two main groups: pixel based and object based change detection. The former are attempts to identify whether or not a change has occurred at each pixel in the image pair, while the latter methods attempt to first group pixels that belong to the same object and use information such as the object's colour, shape and neighbourhood to help determine if that object has been changed between the acquisitions. Change detection algorithms can also be split into supervised and unsupervised groups.

As noted by \citet{hussain2013change} and \citet{bruzzone2013novel}, change detection on low resolution images and on VHR images face different challenges. In low resolution images, pixels frequently contain information about several objects contained within its area. In such cases, a pixel in an image pair may contain both changed and unchanged surfaces simultaneously. VHR images are more susceptible to problems such as parallax, high reflectance variability for objects of the same class, and co-registration problems~\citep{bruzzone2013novel}. It follows that algorithms that perform change detection on very high resolution images must be aware of not only a given pixel's values, but also of information about its neighbourhood. 

Machine learning algorithms, and notably convolutional neural networks (CNNs) in recent years, also have had great impact. For examples, in remote sensing, CNNs were used for road detection~\citep{mnih2010learning}, and in computer vision, CNNs were used on the related task of comparing image pairs~\citep{chopra2005learning, zagoruyko2015learning}. We now examine in details unsupervised and supervised machine learning approaches, the latter category being then subdivided in standard techniques, CNNs and Fully-Convolutional Neural Networks.

\textbf{Unsupervised methods} have been used for change detection in many different ways~\citep{hussain2013change, vakalopoulou2015simultaneous, liu2019contrario}. In the context of change detection, annotated datasets are extremely scarce and often kept private. Thus, unsupervised methods are extremely useful, since, unlike supervised methods, they do not need labelled data for training. Many of these methods automatically analyse the data in difference images and detect patterns that correspond to changes~\citep{bazi2005unsupervised, bruzzone2000automatic}. Other methods use unsupervised learning approaches such as iterative training~\citep{liu2016deep}, autoencoders~\citep{zhao2014deep}, and principal component analysis with $k$-means clustering~\citep{celik2009unsupervised} to separate changed pixels from unchanged ones.

\textbf{Supervised change detection} algorithms require labelled training data from which the task of change detection can be learned. Several methods have been proposed for performing change detection using supervised learning algorithms such as support vector machines~\citep{huang2008use, volpi2009supervised, volpi2013supervised, le2013urban}, random forests~\citep{sesnie2008integrating}, and neural networks~\citep{gopal1996remote, dai1999remotely, zhao2014deep}. CNN architectures have also been proposed to perform supervised change detection~\citep{zhan2017change, chen2018mfcnet}.

Convolutional neural networks (CNNs) for change detection have been proposed by different authors in the recent years. The majority of these methods avoid the problem of the lack of data by using transfer learning techniques, i.e. using networks that have been pre-trained for a different purpose on a large dataset~\citep{el2016convolutional, el2017zoom}. While transfer learning is a valid solution, it is also limiting. Firstly, end-to-end training tends to achieve the best results for a given problem when possible. Transfer learning also assumes that all images are of the same type. As most large scale datasets contain RGB images, this means that extra bands contained in multispectral images must be ignored. It has however been shown that using all available multispectral bands for change detection leads to better results \citep{daudt2018urban}.

Several works have used CNNs to generate the difference image that was described earlier, followed by traditional thresholding methods on those images. \citet{el2016convolutional, el2017zoom} proposed using the activation of pre-trained CNNs to generate descriptors for each pixel, and using the Euclidean distance between these descriptors to build the difference image. \citet{zhan2017change} trained a network to produce a 16-dimensional descriptor for each pixel. Descriptors were similar for pixels with no change and dissimilar for pixels that experienced change. \citet{liu2016deep} used deep belief networks to generate pixel descriptors from heterogeneous image pairs, then the Euclidean distance is used to build a difference image. \citet{zhao2014deep} proposed a deep belief network that takes into account the context of a pixel to build its descriptor. \citet{mou2018learning} proposed using patch based recurrent CNNs to detect changes in image pairs. CNNs for change detection have also been studied outside the context of remote sensing, such as surface inspection \citep{stent2015detecting}.

Fully convolutional neural networks (FCNNs) are a type of CNNs that are especially suited for dense prediction of labels and semantic segmentation~\citep{long2015fully}. Unlike traditional CNNs, which output a single prediction for each input image, FCNNs are able to predict labels for each pixel independently and efficiently. \citet{ronneberger2015u} proposed a simple and elegant addition to FCNNs that aims to improve the accuracy of the final prediction results. The proposed idea is to connect directly layers in earlier stages of the network to layers at later stages to recover accurate spatial information of region boundaries. FCNNs currently achieve state-of-the-art results in semantic segmentation problems, including those in remote sensing~\citep{volpi2017dense, maggiori2017high, chen2018semantic}.

Fully convolutional networks trained from scratch to perform change detection were proposed for the first time by \citet{daudt2018fully}. Both Siamese and early fusion architectures were compared, expanding on the ideas proposed earlier by \citet{chopra2005learning} and \citet{zagoruyko2015learning}. A similar approach was simultaneously proposed by \citet{chen2018mfcnet} outside the context of remote sensing. To the best of our knowledge, the only other time a fully convolutional Siamese network has been proposed was by \citet{bertinetto2016fully} with the purpose of tracking objects in image sequences.

\section{Dataset}\label{sec:dataset}

Research on the problem of change detection is hindered by a lack of open datasets. Such datasets are essential for a methodical evaluation of different algorithms. \citet{benedek2009change} created a binary change dataset with 13 aerial image pairs split into three regions called the Air Change dataset. A dataset, called ONERA Satellite Change Detection (OSCD) dataset, composed of 24 multispectral image pairs taken by the Sentinel-2 satellites is presented in \citep{daudt2018urban}. Both of these datasets allow for simple machine learning techniques to be applied to the problem of change detection, but with these small amounts of images overfitting becomes one of the main concerns even with relatively simple models. The Aerial Imagery Change Detection (AICD) dataset contains synthetic aerial images with artificial changes generated with a rendering engine~\citep{bourdis2011constrained}. These datasets do not contain semantic information about the land cover of the images, and contain either low resolution (OSCD, Air Change) or simulated (AICD) images.

For this reason, we have created the first large scale dataset for semantic change detection, which we present in this section. The High Resolution Semantic Change Detection (HRSCD) dataset will be released to the scientific community to be used as a benchmark for semantic change detection algorithms and to open the doors to the usage of state-of-the-art deep learning algorithms in this context. The dataset contains not only information about where changes have taken place, but also semantic information about the imaged terrain in all images of the dataset. Examples of image pairs, land cover maps (LCM) and change maps taken from the dataset are depicted in Fig.~\ref{fig:example}.

\subsection{Images}

The dataset contains a total of 291 RGB image pairs of 10000x10000 pixels. These are mosaics of aerial images taken by the French National Institute of Geographical and Forest Information (IGN). The image pairs contain an earlier image acquired in 2005 or 2006, and a second image acquired in 2012. They come from a database named \textit{BD ORTHO} which contains orthorectified aerial images of several regions of France from different years at a resolution of 50~cm per pixel. The 291 selected image pairs are all the images in this database that satisfy the conditions for the labels, which will be described below. The images cover a range of urban and countryside areas around the French cities of Rennes and Caen.

The dataset contains more than 3000 times more annotated pixel pairs than either OSCD or Air Change datasets. Also, unlike these datasets, the labels contain information about the types of change that have occurred. Finally, labels about the land cover of the images in the dataset are also available. This is much more data than was previously available in the context of change detection and it opens the doors for many new ideas to be tested. The amount of labelled pixels and surface area for land cover classification is also about 8 times larger in the proposed HRSCD dataset than in the DeepGlobe Land Cover Classification dataset~\citep{deepglobe}, both of the datasets containing images of the same spatial resolution (50~cm/px).

The \textit{BD ORTHO} images provided by IGN are available for free for research purposes, but not all images can be redistributed by the users. That is the case for the images taken in 2005 and 2006. Nevertheless, we will make available all the data for which we have the rights of redistribution and the rasters that we have generated for semantic change detection and land cover mapping. The dataset will also contain instructions for downloading the remaining images that are necessary for using the dataset directly from IGN's website.

\subsection{Labels}

The labels in the dataset come from the European Environment Agency's (EEA) Copernicus Land Monitoring Service - Urban Atlas project. It provides "reliable, inter-comparable, high-resolution land use maps" for functional urban areas in Europe with more than 50000 inhabitants. These maps were generated for the years of 2006 and 2012, and a third map is available containing the changes that took place in that period. Only the images in the regions mapped in the Urban Atlas project and with a maximum temporal distance of one year were kept in the dataset.

The available land cover maps are divided in several semantic classes, which are in turn organised in different hierarchical levels. By grouping the labels at different hierarchical levels it is possible to generate maps that are more coarsely or finely divided. For example, grouping the labels with the coarsest hierarchical level yields five classes (plus the "no information" class) shown in Table \ref{tab:l1-classes}. This hierarchical level will henceforth be referred to as L1.

\begin{table}[t]
    \caption{\label{tab:l1-classes}Urban Atlas land cover mapping classes at hierarchical level L1}
    \centering
    \begin{tabular}{|l|l|}
        \hline
        \textbf{Code} & \textbf{Class} \\
        \hline
        0 & No information \\
        \hline
        1 & Artificial surfaces \\
        \hline
        2 & Agricultural areas \\
        \hline
        3 & Forests \\
        \hline
        4 & Wetlands \\
        \hline
        5 & Water \\
        \hline
    \end{tabular}
\end{table}

These maps are openly available in vector form online. We have used these vector maps and the georeferenced \textit{BD ORTHO} images to generate rasters of the vector maps that are aligned with the rasters of the images. These rasters allow us to have ground truth information about each pixel in the dataset.

It is important to note that there are slight differences in the semantic classes present in Urban Atlas 2006 and in Urban Atlas 2012. These differences do not affect the L1 hierarchical grouping and therefore had no consequence in the work presented later in this paper. It may nevertheless affect future works done with the data. We leave it up to the users how to best interpret and deal with these differences. More information will be provided in the dataset files.

\subsection{Dataset analysis}

Despite its unprecedented size and qualities, we acknowledge in this section the dataset's limitations and challenges. Nevertheless, we will show later in this paper that despite these limitations, the dataset allows for the boundaries of the state-of-the-art in semantic change detection through machine learning to be pushed.

\begin{figure}[!t]
    \centering
    \begin{minipage}[b]{0.3\linewidth}
        \centering
        \centerline{\epsfig{figure=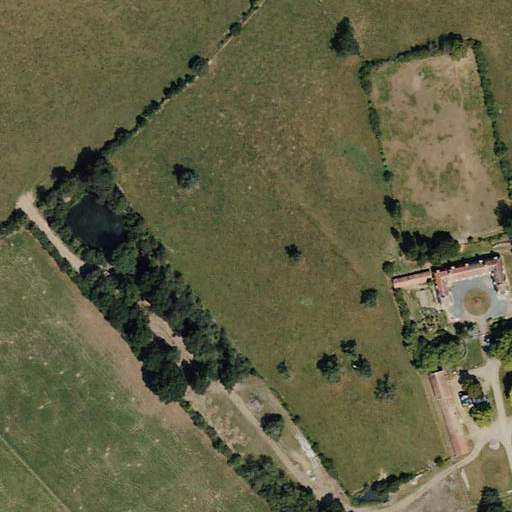,width=0.9\linewidth}}
        \centerline{(a) Image 1.}\medskip
    \end{minipage}
    \hfill
    \begin{minipage}[b]{0.3\linewidth}
        \centering
        \centerline{\epsfig{figure=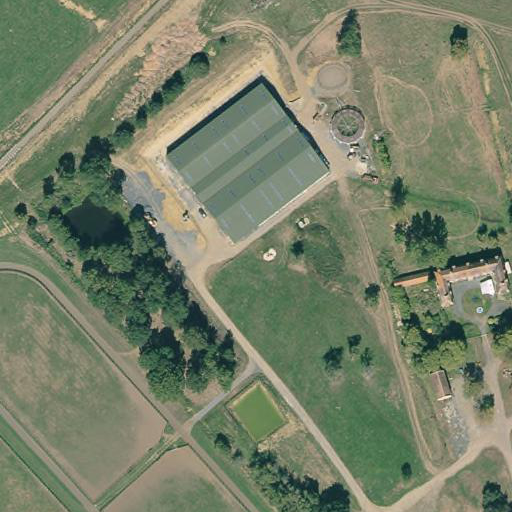,width=0.9\linewidth}}
        \centerline{(b) Image 2.}\medskip
    \end{minipage}
    \hfill
    \begin{minipage}[b]{0.3\linewidth}
        \centering
        \centerline{\epsfig{figure=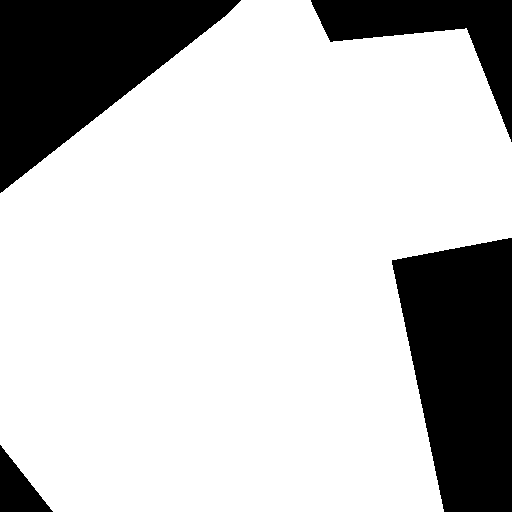,width=0.9\linewidth}}
        \centerline{(c) Inaccurate border.}\medskip
    \end{minipage}
    \\
    \begin{minipage}[b]{0.3\linewidth}
        \centering
        \centerline{\epsfig{figure=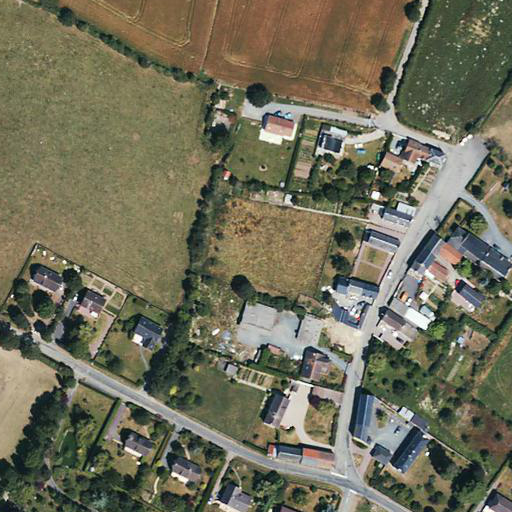,width=0.9\linewidth}}
        \centerline{(d) Image 1.}\medskip
    \end{minipage}
    \hfill
    \begin{minipage}[b]{0.3\linewidth}
        \centering
        \centerline{\epsfig{figure=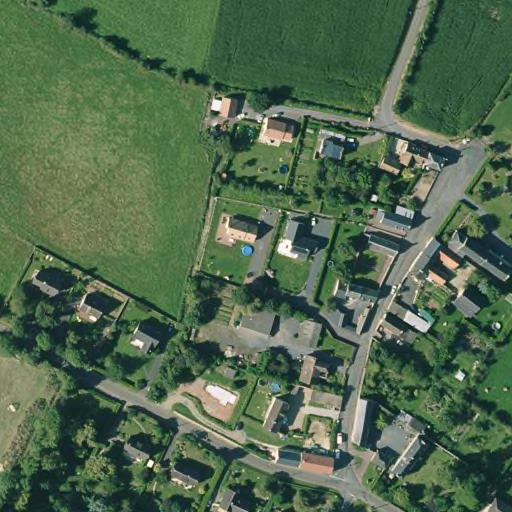,width=0.9\linewidth}}
        \centerline{(e) Image 2.}\medskip
    \end{minipage}
    \hfill
    \begin{minipage}[b]{0.3\linewidth}
        \centering
        \centerline{\epsfig{figure=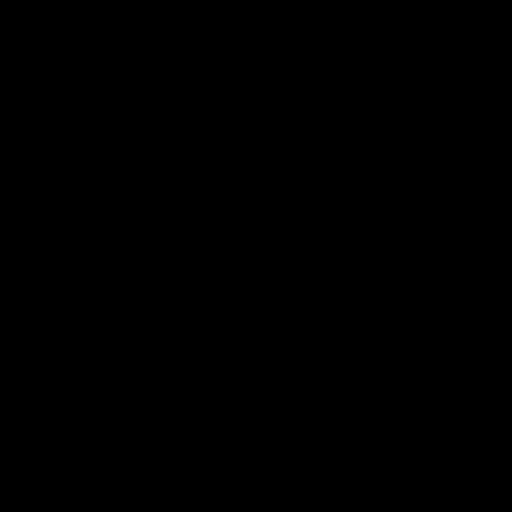,width=0.9\linewidth}}
        \centerline{(f) False negative.}\medskip
    \end{minipage}
    \\
    \begin{minipage}[b]{0.3\linewidth}
        \centering
        \centerline{\epsfig{figure=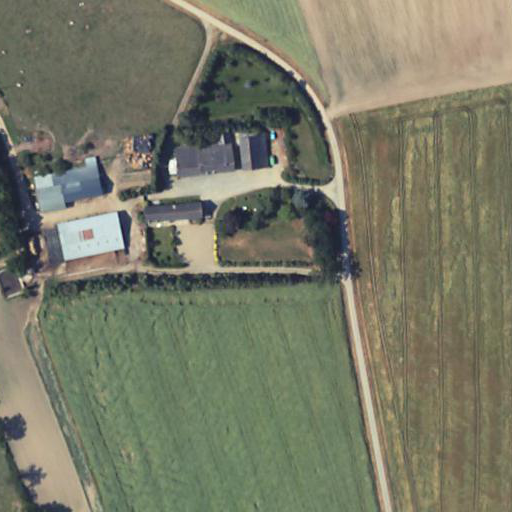,width=0.9\linewidth}}
        \centerline{(g) Image 1.}\medskip
    \end{minipage}
    \hfill
    \begin{minipage}[b]{0.3\linewidth}
        \centering
        \centerline{\epsfig{figure=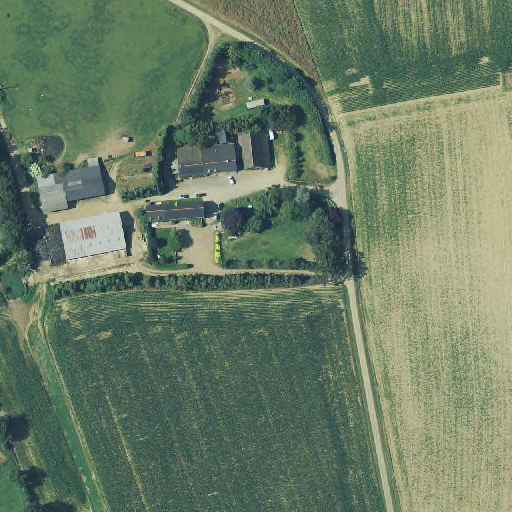,width=0.9\linewidth}}
        \centerline{(h) Image 2.}\medskip
    \end{minipage}
    \hfill
    \begin{minipage}[b]{0.3\linewidth}
        \centering
        \centerline{\epsfig{figure=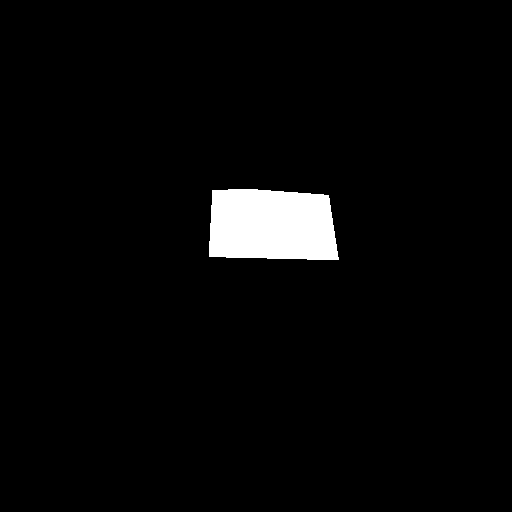,width=0.9\linewidth}}
        \centerline{(i) False positive.}\medskip
    \end{minipage}
    \caption{\label{fig:bad-examples}Examples of: ((a)-(c)) overly large change markings, ((d)-(f)) failure to mark changes, ((g)-(i)) false positive.}
\end{figure}

One issue is the accuracy of the labels contained in the Urban Atlas vector maps with respect to the \textit{BD ORTHO} images. We do not have access to the images used to build the Urban Atlas vector maps, nor to the exact dates of their acquisitions, nor to the dates of acquisition of the images in \textit{BD ORTHO}. Hence, there are some discrepancies between the information in the vector maps and in the images. Furthermore, EEA only guarantees a minimum label accuracy of 80-85\% depending on the considered class. Most of the available data is accurate, but it is important to consider that the labels in the dataset are not flawless. Examples of false negatives and false positives can be see in Fig.~\ref{fig:bad-examples}~(d)-(f) and Fig.~\ref{fig:bad-examples}~(g)-(i), respectively.
It is also worth noting that the labels have been created using previously known vector maps, mostly by labelling correctly each of the known regions. This means a single label was given to each region, and this led to inaccurate borders in some cases. This can be clearly seen in Fig.~\ref{fig:bad-examples}~(a)-(c).

One of the main challenges involved in using this dataset for supervised learning is the extreme label imbalance. As can be seen in Table~\ref{tab:l1-imbalance}, 99.232\% of all pixels are labelled as no change, and the largest class is from agricultural areas to artificial surfaces (i.e. class 2 to class 1), which accounts for 0.653\% of all pixels. These two classes together account for 99.885\% of all pixels, which means all other change types combined account for only 0.115\% of all pixels. Furthermore, many of the possible types of change have no examples at all in any of the images of the dataset. It is of paramount importance when using this dataset to take into account this imbalance. This also means that using the overall accuracy as a performance metric with this dataset is not a good choice, as it virtually only reflects how many pixels of the no change class have been classified correctly. Other metrics, such as Cohen's kappa coefficient or the S{\o}rensen-Dice coefficient, must be used instead. This class imbalance is characteristic of real world large scale data, where changes are much less frequent than unchanged surfaces. Therefore, this dataset provides a realistic evaluation tool for change detection methods, unlike carefully selected image pairs with large changed regions.

\begin{table}[t]
    \caption{\label{tab:l1-imbalance}Change class imbalance at hierarchical level L1. Row number represents class in 2006, column number represents class in 2012. Classes were defined in Table~\ref{tab:l1-classes}.}
    \centering
    \begin{tabular}{|c|c|c|c|c|c|}
        \hline
         & 1 & 2 & 3 & 4 & 5 \\
        \hline
        1 & 0\% & 0.011\% & 0\% & 0.001\% & 0.001\% \\
        \hline
        2 & 0.653\% & 0\% & 0.001\% & 0\% & 0.077\% \\
        \hline
        3 & 0.014\% &  0.002\% & 0\% & 0\% & 0\% \\
        \hline
        4 & 0\% & 0\% & 0\% & 0\% & 0\% \\
        \hline
        5 & 0.001\% & 0.004\% & 0\% & 0.004\% & 0\% \\
        \hline
        \hline
        \multicolumn{3}{|c|}{No change} &  \multicolumn{3}{|c|}{99.232\%}  \\
        \hline
    \end{tabular}
\end{table}

The problem of supervised learning using noisy labels has already been studied and evidence suggests that supervised learning with noisy labels is possible as long as a dataset of a large enough size is used \citep{rolnick2017deep}. Other works attempt to explicitly deal with the noisy labels present in the dataset and prioritise the correct labels during training \citep{maggiolo2018improving}.

Finally, we acknowledge how challenging it is to use hierarchical levels finer than L1 due to: 1) a massive increase in the number of possible changes, and 2) the difference between similar classes becomes more abstract and context based. For example, the difference between the "Discontinuous Medium Density Urban Fabric" and the "Discontinuous Low Density Urban Fabric" classes defined in Urban Atlas depends not only in correctly identifying the surface at a given pixel (e.g. building or grass), but also by understanding the surroundings of the pixel and calculating the ratio between these two classes at a given neighbourhood that is not clearly defined.

\section{Methodology}

\subsection{Binary change detection}
\label{sec:binary}

We have already showed in a previous work the efficacy of using three different architectures of fully convolutional neural networks for change detection \citep{daudt2018fully}. \citet{chen2018mfcnet} simultaneously proposed a fully convolutional architecture for change detection that is very similar to one of the three initially proposed architectures. In both of these works, FCNN architectures performed better than previous methods for change detection.

\begin{figure}[!ht]
    \centering
    \begin{minipage}[b]{0.75\linewidth}
        \centering
        \centerline{\epsfig{figure=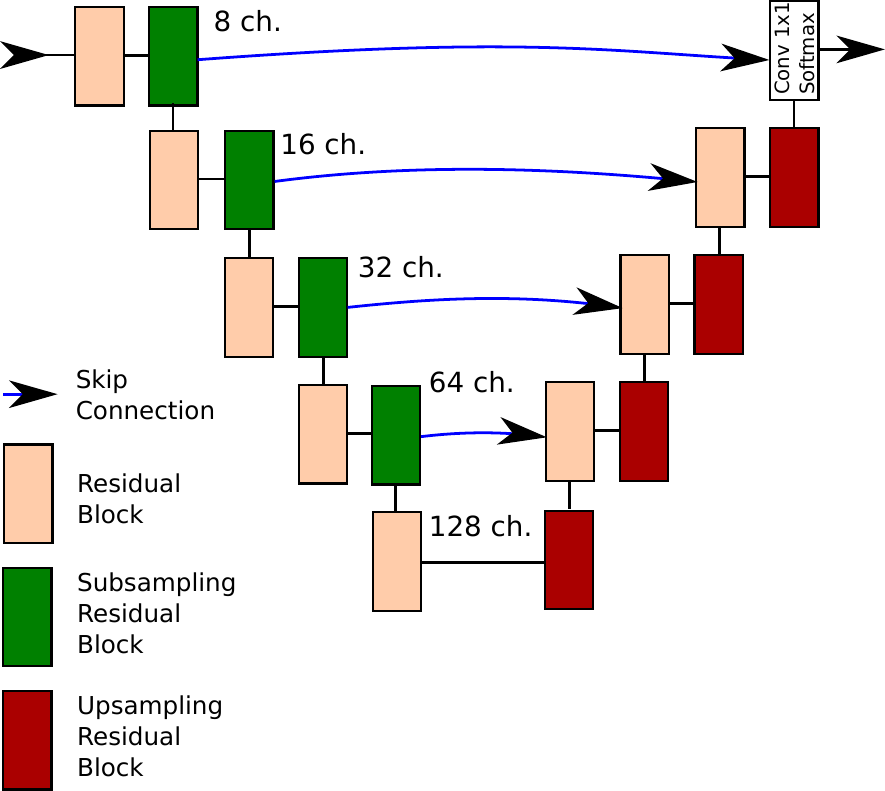,width=\linewidth}}
    \end{minipage}
    
    \caption{\label{fig:fresunet}FC-EF-Res architecture, used for tests with smaller datasets to avoid overfitting. Using residual blocks improves network performance and facilitates training.}
\end{figure}

Building on this previous work, we have modified the FC-EF architecture proposed in \citet{daudt2018fully} to use residual blocks, as proposed by \citet{he2016deep}. The resulting network is later referred to as FC-EF-Res, and is depicted in Fig.~\ref{fig:fresunet}. These residual blocks were used in an encoder-decoder architecture with skip connections to improve the spatial accuracy of the results \citep{ronneberger2015u}. These residual blocks were chosen to facilitate the training of the network, which is especially important for its deeper variations that will be discussed later.

When testing on the OSCD dataset (Section~\ref{sec:oscd}), the size of the network has been kept approximately the same as in \citet{daudt2018fully} to avoid overfitting. 
When using the proposed HRSCD dataset (Section~\ref{sec:hrscd}), the larger amount of annotated pixels allows us to use deeper and more complex models. In that case, the number of encoding levels and residual blocks per level has been increased, but the idea behind the network is the same as of FC-EF-Res.

\subsection{Semantic change detection}\label{sec:strategies}

As was mentioned earlier, the efficiency of the proposed architecture for binary change detection and the availability of the HRSCD dataset enable us to tackle the problem of semantic change detection. This problem consists of two separate but not independent parts. The first task is analogue to binary change detection, i.e. we attempt to determine whether a change has occurred at each pixel in a co-registered multi-temporal image pair. The second task is to differentiate between types of changes. In our case, this consists of predicting the class of the pixel in each of the two given images. The problem of semantic change detection lies in the intersection between change detection and land cover mapping.

Below we will describe four different intuitive strategies to perform semantic change detection using deep neural networks. Starting from the plain comparison of land cover maps, we then develop more involved strategies. These strategies vary in complexity and performance, as will be discussed in Section~\ref{sec:results}.

\begin{figure*}[!ht]
    \centering
    \hfill
    \begin{minipage}[b]{0.4\linewidth}
        \centering
        \centerline{\epsfig{figure=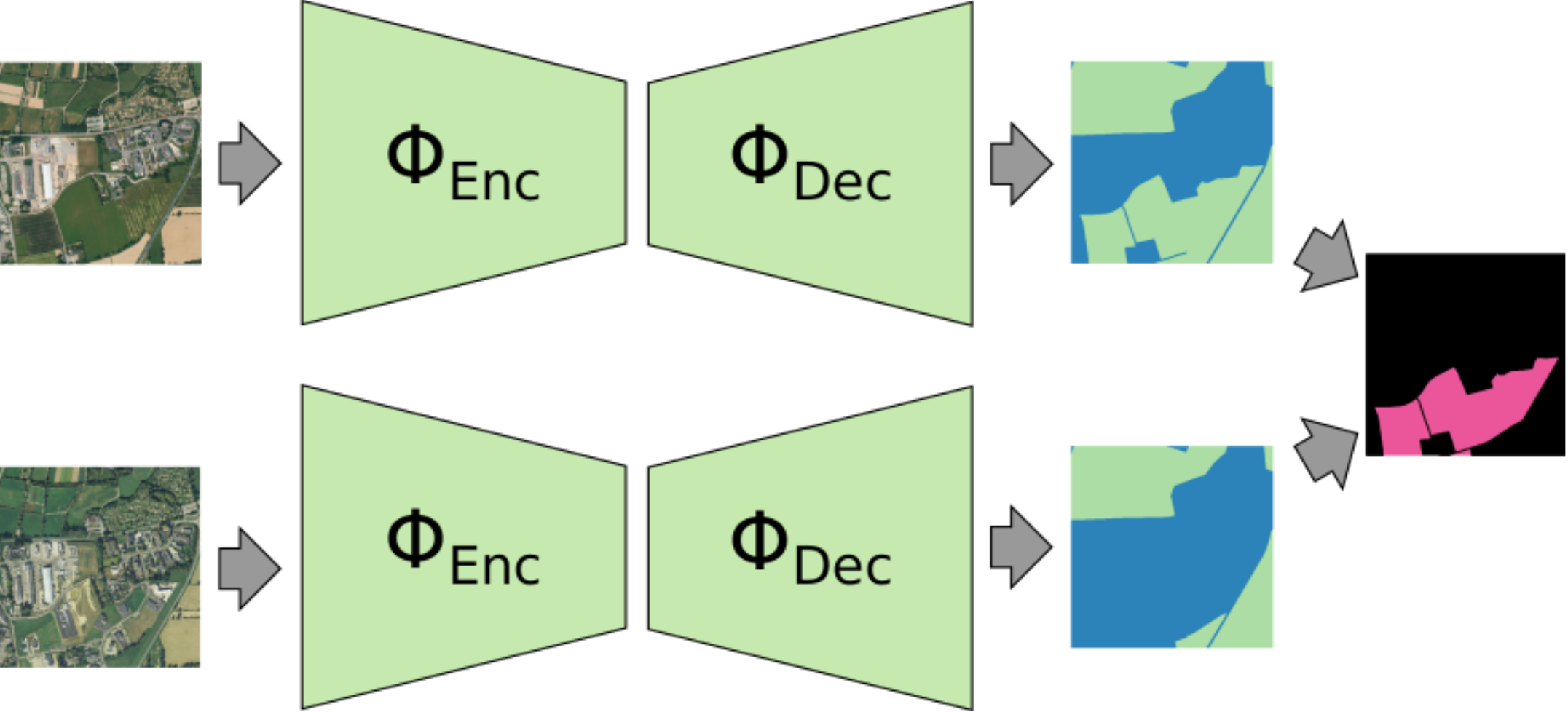,scale=0.3}}
        \centerline{(a) Strategy 1: semantic CD from land cover maps.}\medskip
    \end{minipage}
    \hfill
    \begin{minipage}[b]{0.4\linewidth}
        \centering
        \centerline{\epsfig{figure=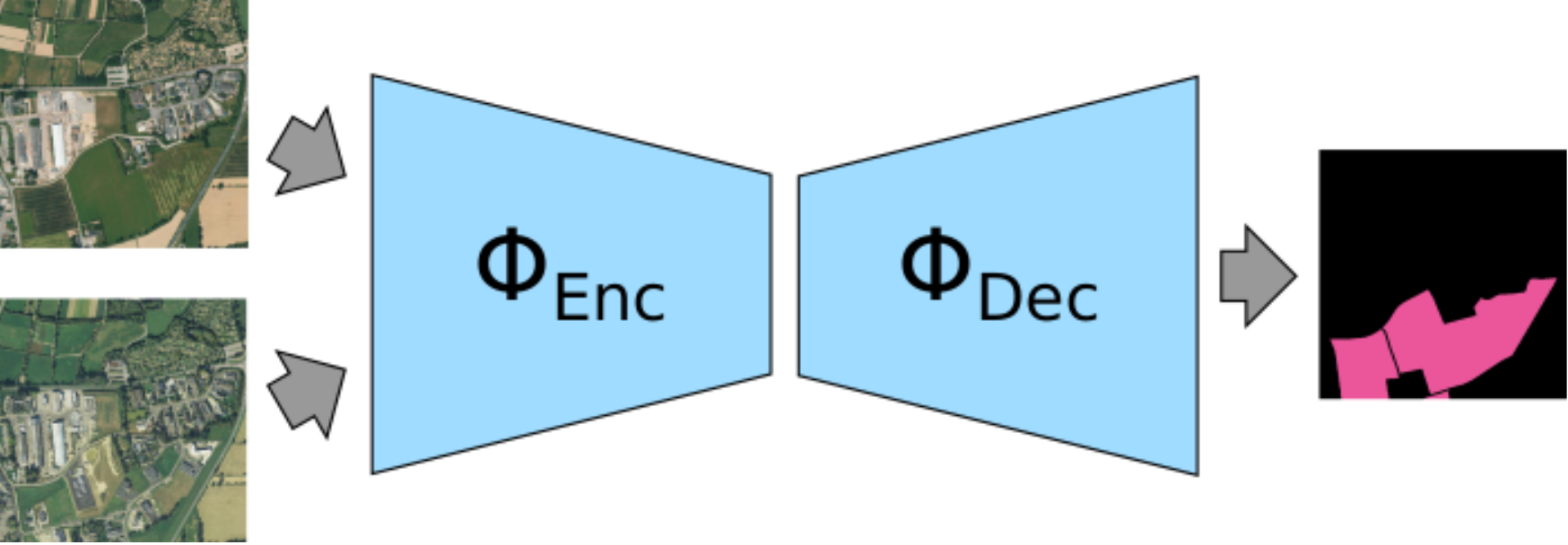,scale=0.3}}
        \centerline{(b) Strategy 2: direct semantic CD.}\medskip
    \end{minipage}
    \hfill
    \begin{minipage}[b]{0.1\linewidth}
    \end{minipage}
    \\
    \hfill
    \begin{minipage}[b]{0.4\linewidth}
        \centering
        \centerline{\epsfig{figure=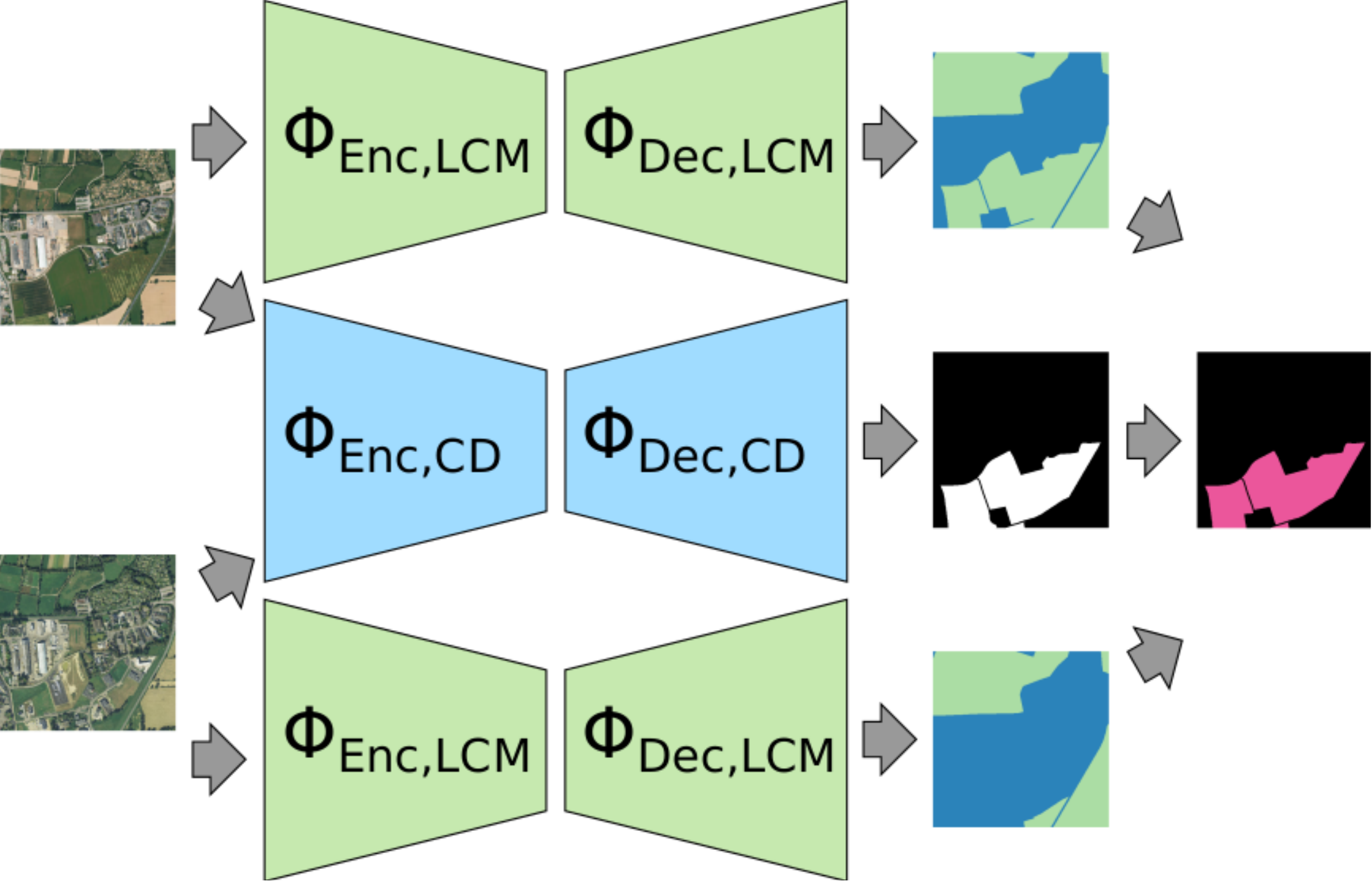,scale=0.3}}
        \centerline{(c) Strategy 3: separate CD and LCM.}\medskip
    \end{minipage}
    \hfill
    \begin{minipage}[b]{0.4\linewidth}
        \centering
        \centerline{\epsfig{figure=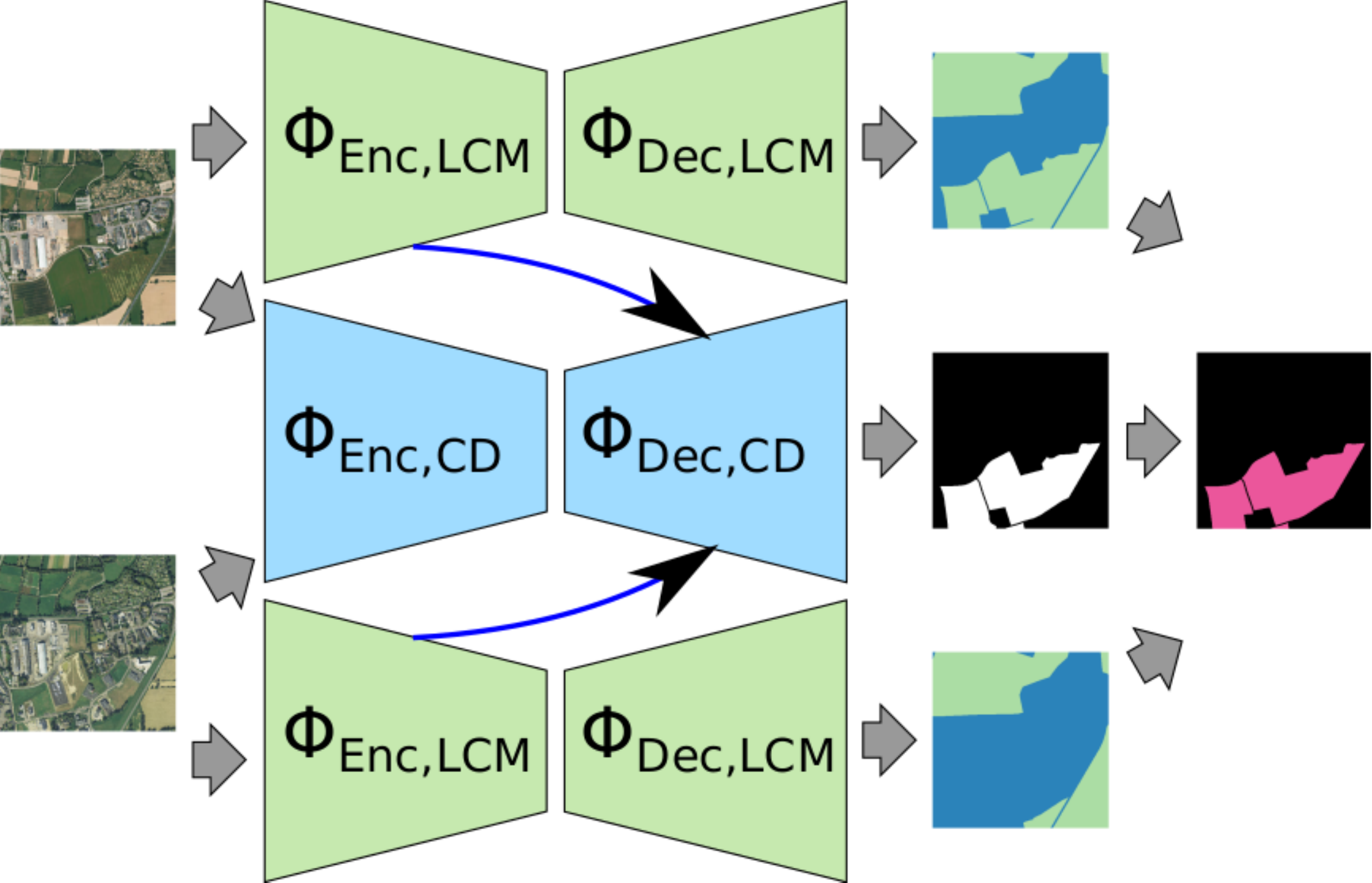,scale=0.3}}
        \centerline{(d) Strategy 4: integrated CD and LCM.}\medskip
    \end{minipage}
    \hfill
    \begin{minipage}[b]{0.1\linewidth}
    \end{minipage}
    
    \caption{\label{fig:strategies}Schematics for all four proposed strategies for semantic change detection. $\Phi$ represents the network branch's learnable parameters, "Enc" means encoder, "Dec" means decoder, "LCM" means land cover mapping, and "CD" means change detection.}
\end{figure*}

\subsubsection{Strategy 1: Direct comparison of LCMs}

The problem of automatic land cover mapping is a well studied problem. In particular, methods involving CNNs have recently been proposed, yielding good performances \citep{audebert2016semantic}. When the land cover information is available, as it is the case in the HRSCD dataset, the most intuitive method that can be proposed for semantic change detection would be to train a land cover mapping network and to compare the results for pixels in the image pair (see Fig.~\ref{fig:strategies}(a)).

The advantage of this method is its simplicity. In many cases we could assume changes occurred where the predicted class label differs between the two images, and the type of change is given by the predicted labels at each of the two acquisition moments. The weakness of this method is that it heavily depends on the accuracy of the predicted land cover maps. While modern FCNNs are able to map areas to a good degree of accuracy, there are still many wrongly predicted labels, especially around the boundaries between regions of different classes. Furthermore, when comparing the results for two acquisitions the prediction errors would accumulate. This means the accuracy of this change detection algorithm would be lower than the land cover mapping network, and would likely predict changes in the borders between classes simply due to the inaccuracy of the network. 

\subsubsection{Strategy 2: Direct semantic CD}

A second intuitive approach is to treat each possible type of change as a different and independent label, considering semantic change detection as a simple semantic segmentation along the lines of what has been done to binary change detection in the past \citep{daudt2018fully}.

The weakness of this method is that the number of change classes grows proportionately to the square of the number of land cover classes that is considered. This, combined with the class imbalance problem that was discussed earlier, proves to be a major challenge when training the network.

\subsubsection{Strategy 3: Separate LCM and CD}

Since it has been proven before that FCNNs are able to perform both binary change detection and land cover mapping, a third possible approach is to train two separate networks that together perform semantic change detection (see Fig.~\ref{fig:strategies}(c)). The first network performs binary change detection on the image pair, while the second network performs land cover mapping of each of the input images. The two networks are trained separately since they are independent.

In this strategy, the two input images produce three outputs: two land cover maps and a change map. At each pixel, the presence of change is predicted by the change map, and the type of change is defined by the classes predicted by the land cover maps at that location. This way the number of predicted classes is reduced relative to the previous strategy (i.e. the number of classes is no longer proportional to the square of land cover classes) without loss of flexibility. This helps with the class imbalance problem. It also avoids the problem of predicting changes at every pixel where the land cover maps differ, since the change detection problem is treated separately from land cover mapping.

We argue that such network may be able to identify changes of types it has not seen during training, as long as it has seen the land cover classes during training. For example, the network could in theory correctly classify a change from agricultural area to wetland even if such changes are not in the training set, as long as it has enough examples of those classes to correctly classify them in the land cover mapping branches. The combination of two separate networks allows us to split the problem into two, and optimise each part to maximise performance.

\begin{table}[t]
    \caption{\label{tab:strategies}Summary of proposed change detection strategies.}
    \centering
    \begin{tabular}{|c|c|c|}
        \hline
        Str. & Description & Training \\
        \hline \hline
        1 & Diff. of LCMs & LCM supervision \\  \hline
        2 & Direct semantic CD & Multiclass CD supervision \\  \hline
        3 & Separate CD and LCM & Separate LCM and CD \\  \hline
        4.1 & Integrated CD and LCM & Triple loss function \\  \hline
        4.2 & Integrated CD and LCM & Sequential training \\  \hline
    \end{tabular}
\end{table}

\subsubsection{Strategy 4: Integrated LCM and CD}

\begin{figure*}[!ht]
    \centering
    \begin{minipage}[b]{0.9\linewidth}
        \centering
        \centerline{\epsfig{figure=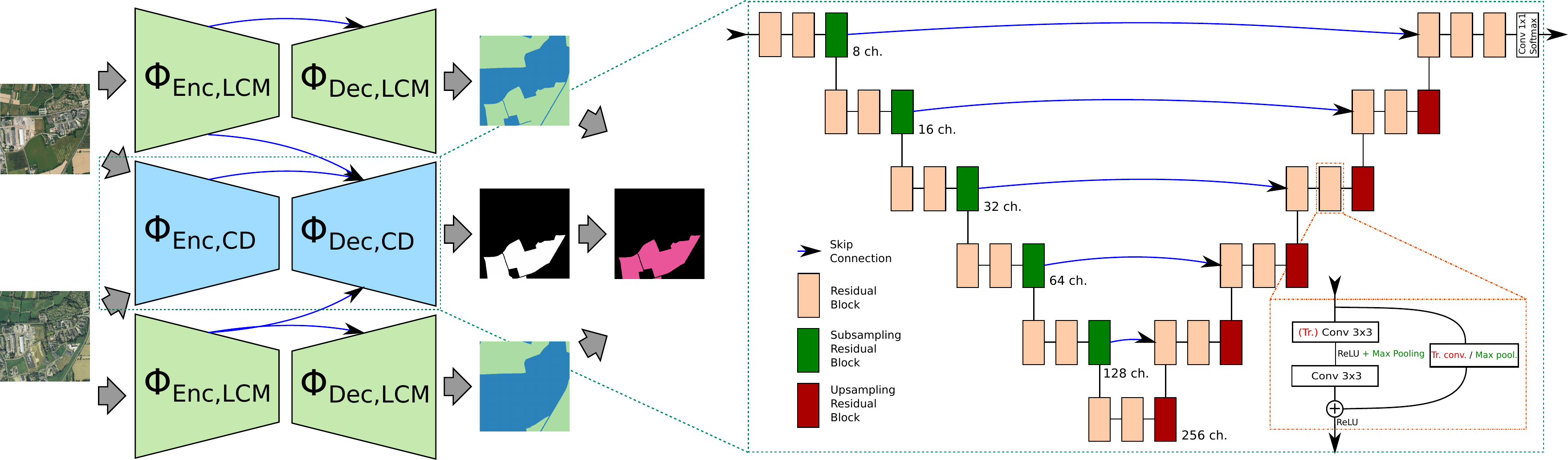,width=\linewidth}}
    \end{minipage}
    
    \caption{\label{fig:frankennet}Detailed schematics for the integrated change detection and land cover mapping network (Strategy 4). The encoder-decoder architecture is the same that was used for all 4 strategies.}
\end{figure*}

The last of the proposed approaches is an evolution of the previous strategy of using two FCNNs for the tasks of binary change detection and land cover mapping. We propose to integrate the two FCNNs into a single multitask network (see Fig.~\ref{fig:strategies}(d) and Fig.~\ref{fig:frankennet}) so that land cover information can be used for change detection. The combined network takes as input the two co-registered images and outputs three maps: the binary change map and the two land cover maps.

In the proposed architecture, information from the land cover mapping branches of the network is passed to the change detection branch of the network in the form of difference skip connections, which was shown to be the most effective form of skip connections for Siamese FCNNs \citep{daudt2018fully}. The weights of the two land cover mapping branches are shared since they perform an identical task, allowing us to significantly reduce the number of learned parameters.

This multipurpose network gives rise to a new issue during the training phase. Given that the network outputs three different image predictions, it is necessary to balance the loss functions from these results. Since two of the outputs have exactly the same nature (the land cover maps), it follows from the symmetry of these branches that they can be combined into a single loss function by simple addition. The question remains on how to balance the binary change detection loss function and the land cover mapping loss function to maximise performance.

We have proposed and tested two different strategies for training the network. The first and more naive approach to this problem is to minimise a loss function that is a weighted combination of the two loss functions. This loss function would have the form
\begin{equation}
    \begin{split}
        \mathcal{L}_{\lambda}(\Phi_{\rm Enc,CD}, &\Phi_{\rm Dec,CD}, \Phi_{\rm Enc,LCM}, \Phi_{\rm Dec,LCM})\\
        = &\mathcal{L}(\Phi_{\rm Enc,CD}, \Phi_{\rm Dec,CD}) + \lambda  \mathcal{L}(\Phi_{\rm Enc,LCM}, \Phi_{\rm Dec,LCM})
    \end{split}
\end{equation}
where $\Phi$ represents the various network branch parameters, and $\mathcal{L}$ is a pixel-wise loss function. In this work, the pixel-wise cross entropy function was used as loss function as is traditional in semantic segmentation problems. The problem then becomes the search for the value of $\lambda$ that leads to the best balance between the two loss terms. This can be found through a grid search, but the test of each value of $\lambda$ is done by training the whole network until convergence, which is a slow and costly procedure. This will later be referred to as Strategy 4.1.

\begin{table}[t]
    \caption{\label{tab:definitions}Definitions of metrics used for evaluating results quantitatively. Legend: TP - true positive, TN - true negative, FP - false positive, FN - false negative, $p_o$ - observed agreement between ground truth and predictions, $p_e$ - expected agreement between ground truth and predictions given class distributions.}
    \centering
    \begin{tabular}{|c|c|c|c||c|c|}
        \hline
        Tot. acc. & $(TP+TN)/(TP+TN+FP+FN)$ \\    \hline
        Precision & $TP/(TP+FP)$ \\    \hline
        Recall & $TP/(TP+FN)$ \\    \hline
        Dice & $2\cdot TP / (2\cdot TP + FP + FN)$ \\    \hline
        Kappa & $(p_o - p_e)/(1 - p_e)$ \\    \hline
    \end{tabular}
\end{table}

To reduce the aforementioned training burden, we propose a second approach to train the network that avoids the need of setting the hyperparameter $\lambda$. We train the network in two stages. First, we consider only the land cover mapping loss
\begin{equation}
    \begin{split}
        \mathcal{L}_1(\Phi_{\rm Enc,CD}, &\Phi_{\rm Dec,CD}, \Phi_{\rm Enc,LCM}, \Phi_{\rm Dec,LCM})\\
        = &\mathcal{L}(\Phi_{\rm Enc,LCM}, \Phi_{\rm Dec,LCM})
    \end{split}
\end{equation}
and train only the land cover mapping branches of the network, i.e. we do not train $\Phi_{\rm Enc,CD}$ or $\Phi_{\rm Dec,CD}$ at this stage. Since the change detection branch has no influence on the land cover mapping branches, we can train these branches to achieve the maximum possible land cover mapping performance with the given architecture and data. Next, we use a second loss function based only on the change detection branch:
\begin{equation}
    \begin{split}
        \mathcal{L}_2(\Phi_{\rm Enc,CD}, &\Phi_{\rm Dec,CD}, \Phi_{\rm Enc,LCM}, \Phi_{\rm Dec,LCM})\\
        = &\mathcal{L}(\Phi_{\rm Enc,CD}, \Phi_{\rm Dec,CD})
    \end{split}
\end{equation}
while keeping the weights for the land cover mapping $\Phi_{\rm Enc,LCM}$ and $\Phi_{\rm Enc,LCM}$ fixed. This way, the change detection branch learns to use the predicted land cover information to help to detect changes without affecting land cover mapping performance. This will later be referred to as Strategy 4.2.

\section{Results}\label{sec:results}

\subsection{Multispectral change detection}
\label{sec:oscd}

We first evaluate the performance of the proposed FC-EF-Res network. As explained in Section~\ref{sec:binary}, this network is an evolution of the convolutional architecture FC-EF proposed in \citet{daudt2018fully}, to which residual blocks have been added in place of traditional convolutional layers.

The FC-EF-Res architecture was compared to the previously proposed FCNN architectures on the OSCD dataset for binary change detection, which contains lower-resolution Sentinel-2 image pairs with 13 multispectral bands. As expected, the residual extension of the FC-EF architecture outperformed all previously proposed architectures. The difference was noted on both the RGB and the multispectral cases. On the RGB case, the improvement was of such magnitude that the change detection performance on RGB images almost matched the performance on multispectral images. The results can be seen in Table~\ref{tab:oscd-results}. This corroborates the claims made by \citet{he2016deep} that using residual blocks improves the training performance of CNNs. For this reason, all networks that are tested with the HRSCD dataset use residual modules.

\begin{table*}[t]
    \caption{\label{tab:oscd-results}Change detection results of several methods on the OSCD dataset, for the RGB and multispectral (MS) cases. Results are in percent.}
    \centering
    \begin{tabular}{|cc|c|c|c|c|}
        \hline
        Data & Network & Prec. & Recall & Tot. acc. & Dice\\
        \hline \hline
         \parbox[t]{2mm}{\multirow{5}{*}{\rotatebox[origin=c]{90}{RGB}}} & FC-EF        & 44.72 & 53.92 & 94.23 & 48.89 \\ 
                 & FC-Siam-conc & 42.89 & 47.77 & 94.07 & 45.20 \\ 
                 & FC-Siam-diff & 49.81 & 47.94 & 94.86 & 48.86 \\  \cline{2-6}
                 & FC-EF-Res & \textbf{52.27} & \textbf{68.24} & \textbf{95.34} & \textbf{59.20} \\
        \hline \hline 
        \parbox[t]{2mm}{\multirow{5}{*}{\rotatebox[origin=c]{90}{MS}}}   & FC-EF        & \textbf{64.42} & 50.97 & \textbf{96.05} & 56.91 \\ 
                 & FC-Siam-conc & 42.39 & 65.15 & 93.68 & 51.36 \\ 
                 & FC-Siam-diff & 57.84 & 57.99 & 95.68 & 57.92 \\  \cline{2-6}
                 & FC-EF-Res & 54.93 & \textbf{66.48} & 95.64 & \textbf{60.15} \\  \hline
    \end{tabular}
\end{table*}

\subsection{Very high resolution semantic change detection}
\label{sec:hrscd}

\begin{figure*}[!ht]
    \centering
    \begin{minipage}[b]{0.1\linewidth}
        \centering
        \centerline{\epsfig{figure=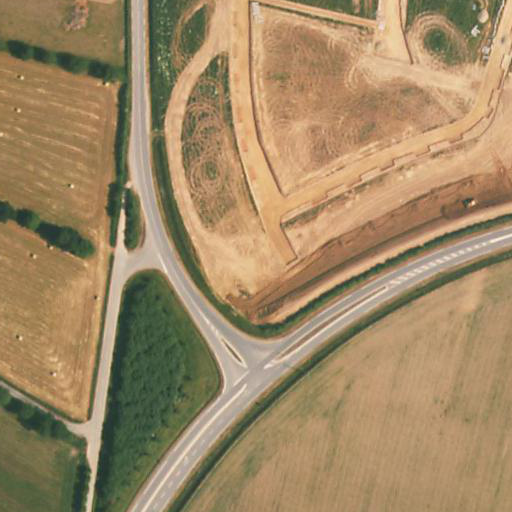,width=\textwidth}}
        \centerline{(a) Image 1}\medskip
    \end{minipage}
    \hfill
    \begin{minipage}[b]{0.1\linewidth}
        \centering
        \centerline{\epsfig{figure=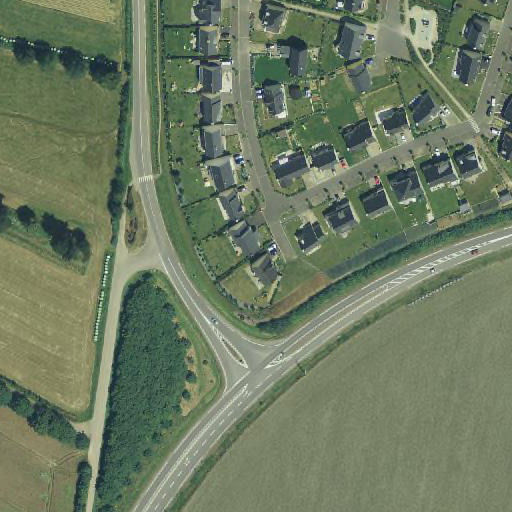,width=\textwidth}}
        \centerline{(b) Image 2}\medskip
    \end{minipage}
    \hfill
    \begin{minipage}[b]{0.1\linewidth}
        \centering
        \centerline{\epsfig{figure=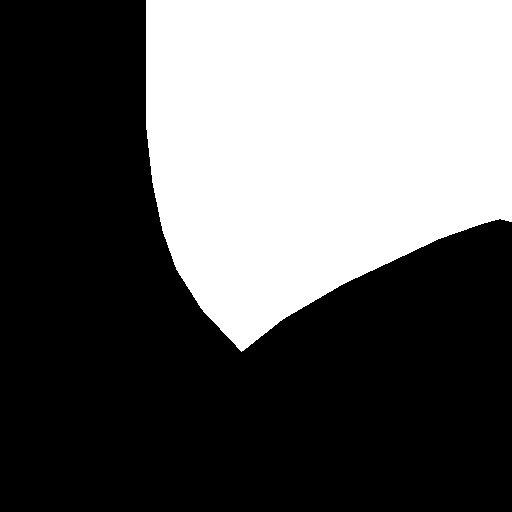,width=\textwidth}}
        \centerline{(c) CD - GT}\medskip
    \end{minipage}
    \hfill
    \begin{minipage}[b]{0.1\linewidth}
        \centering
        \centerline{\epsfig{figure=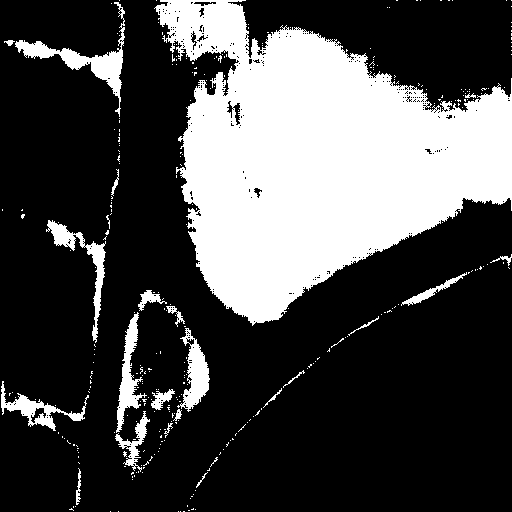,width=\textwidth}}
        \centerline{(d) Str. 1}\medskip
    \end{minipage}
    \hfill
    \begin{minipage}[b]{0.1\linewidth}
        \centering
        \centerline{\epsfig{figure=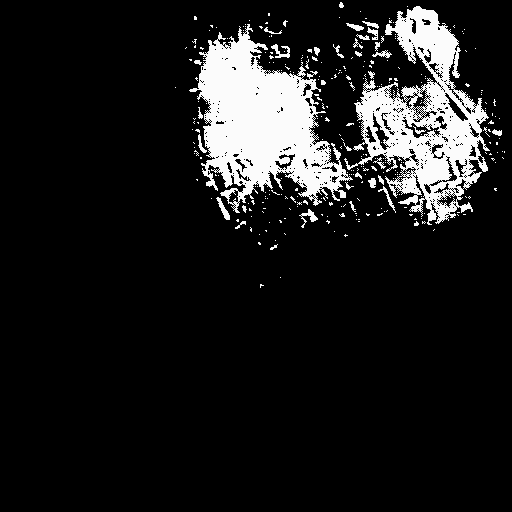,width=\textwidth}}
        \centerline{(e) Str. 2}\medskip
    \end{minipage}
    \hfill
    \begin{minipage}[b]{0.1\linewidth}
        \centering
        \centerline{\epsfig{figure=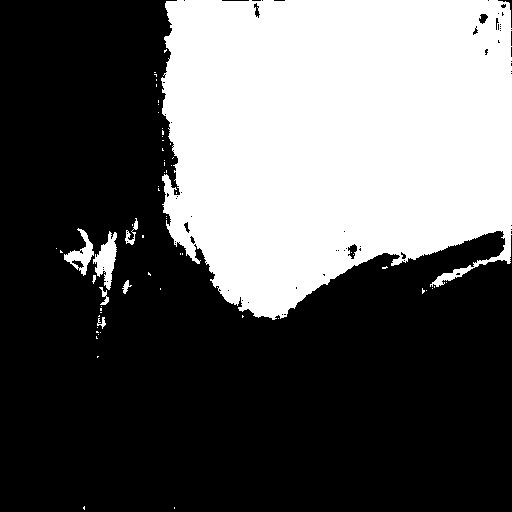,width=\textwidth}}
        \centerline{(f) Str. 3}\medskip
    \end{minipage}
    \hfill
    \begin{minipage}[b]{0.1\linewidth}
        \centering
        \centerline{\epsfig{figure=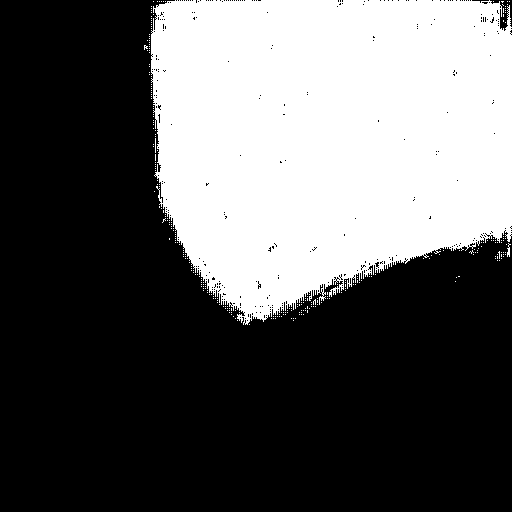,width=\textwidth}}
        \centerline{(g) Str. 4.1}\medskip
    \end{minipage}
    \hfill
    \begin{minipage}[b]{0.1\linewidth}
        \centering
        \centerline{\epsfig{figure=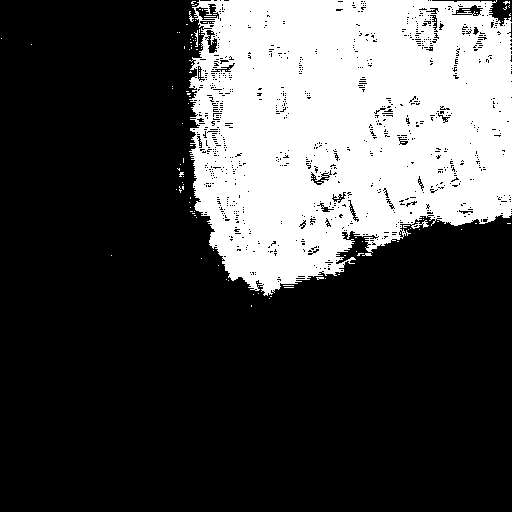,width=\textwidth}}
        \centerline{(h) Str. 4.2}\medskip
    \end{minipage}
    \\
    \begin{minipage}[b]{0.1\linewidth}
        \centering
        \centerline{\epsfig{figure=results-lcm1-gt.png,width=\textwidth}}
        \centerline{(i) LCM 1 - GT}\medskip
    \end{minipage}
    \hfill
    \begin{minipage}[b]{0.1\linewidth}
        \centering
        \centerline{\epsfig{figure=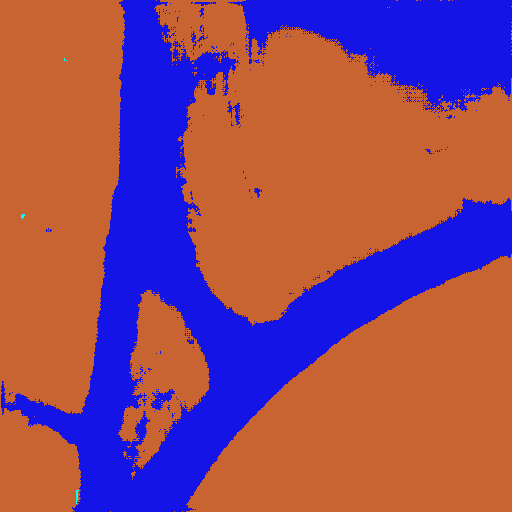,width=\textwidth}}
        \centerline{(j) Str. 1/3}\medskip
    \end{minipage}
    \hfill
    \begin{minipage}[b]{0.1\linewidth}
        \centering
        \centerline{\epsfig{figure=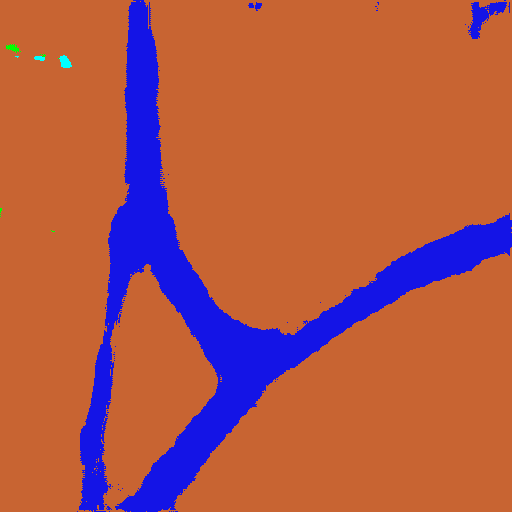,width=\textwidth}}
        \centerline{(k) Str. 4.1}\medskip
    \end{minipage}
    \hfill
    \begin{minipage}[b]{0.1\linewidth}
        \centering
        \centerline{\epsfig{figure=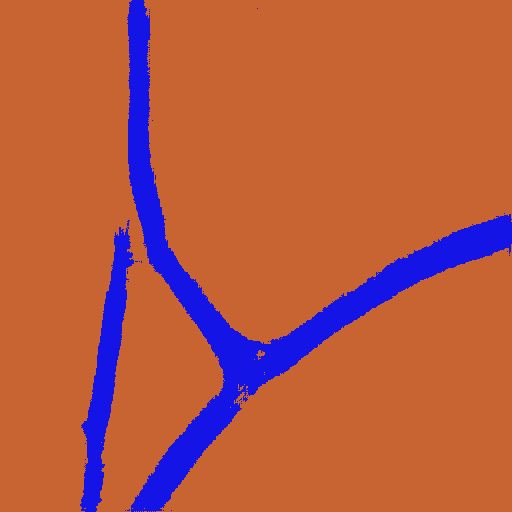,width=\textwidth}}
        \centerline{(l) Str. 4.2}\medskip
    \end{minipage}
    \hfill
    \begin{minipage}[b]{0.1\linewidth}
        \centering
        \centerline{\epsfig{figure=results-lcm2-gt.png,width=\textwidth}}
        \centerline{(m) LCM 2 - GT}\medskip
    \end{minipage}
    \hfill
    \begin{minipage}[b]{0.1\linewidth}
        \centering
        \centerline{\epsfig{figure=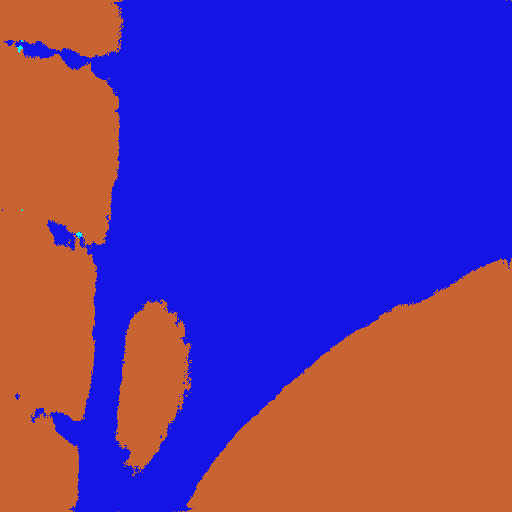,width=\textwidth}}
        \centerline{(n) Str. 1/3}\medskip
    \end{minipage}
    \hfill
    \begin{minipage}[b]{0.1\linewidth}
        \centering
        \centerline{\epsfig{figure=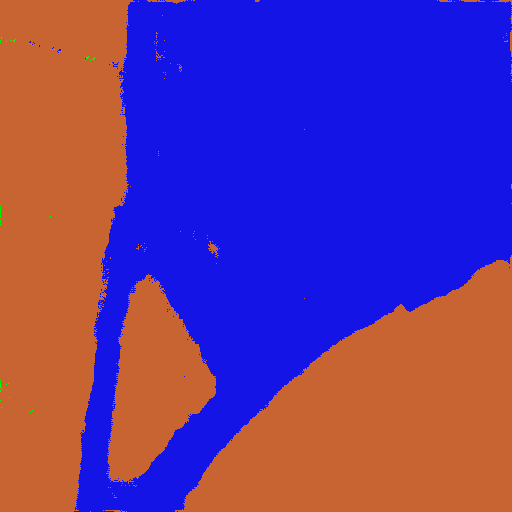,width=\textwidth}}
        \centerline{(o) Str. 4.1}\medskip
    \end{minipage}
    \hfill
    \begin{minipage}[b]{0.1\linewidth}
        \centering
        \centerline{\epsfig{figure=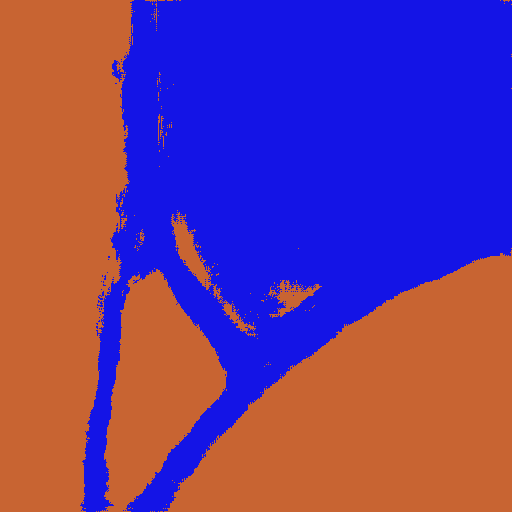,width=\textwidth}}
        \centerline{(p) Str. 4.2}\medskip
    \end{minipage}
    
    \caption{\label{fig:results} Illustrative images of the obtained results: (a)-(b) multitemporal image pair; (c) ground truth change detection map; (d)-(h) predicted change maps; (i)-(l) ground truth and predicted land cover maps for image 1; (m)-(p) ground truth and predicted land cover maps for image 2.}
\end{figure*}

To test the methods proposed in Section~\ref{sec:strategies} we split the HRSCD images into two groups: 146 image pairs for training and 145 image pairs for testing. By splitting the train and test sets this way we can ensure that no pixel in the test set has been seen during training. Class weights were set inversely proportional to the number of training examples to counterbalance the dataset's class imbalance. The results for each of the proposed strategies can be seen in Table~\ref{tab:hrscd-results}, and illustrative image results can be seen in Fig.~\ref{fig:results}.

As is the case for most deep neural networks, the training times for the proposed methods are significantly larger than the testing times. Once the network has been trained, its fast inference speed allows it to process large amounts of data efficiently. The proposed methods took 3-5 hours of training time using a GeForce GTX 1080 Ti GPU with 11GB of memory. Inference times of the proposed methods were under 0.04~s for 512x512 image pairs using the same hardware.

In Strategy 1, which naively attempts to predict change maps from land cover maps, we can see that the network succeeds in accurately classifying the imaged terrains, but this is not enough to predict accurate change maps. The change detection kappa coefficient for this strategy is very low, which means this method is marginally better than chance for change detection.

The results for Strategy 2 are a fair improvement over those of Strategy 1. The change detection Dice coefficient and the land cover mapping results for this method are not reported due to its nature, since Dice coefficients can only be calculated for binary classification problems, and this strategy bypasses the land cover mapping steps. Despite achieving a higher kappa coefficient, the network learned to always predict the same type of change where changes occurred. This means that despite using appropriately tuned class weights, the learning process did not succeed in overcoming the extreme class imbalance present in the dataset. In other words, the network learned to detect changes but no semantic information was present in the results.

For Strategy 3, the land cover mapping network that was used was the same as that of Strategy 1, which achieved good performance. A binary change detection network was trained to be used for masking the land cover maps. The performance of this network was better than that of Strategy 1 but worse than that of Strategy 2. The results show that this is due to an overestimation of the change class. This shows once again how challenging dealing with the extreme class imbalance is.

The results of Strategy 4 are the best ones overall. The simultaneous training strategy (Str. 4.1) achieves excellent performance in both land cover mapping and change detection, proving the viability of this strategy. The reported results were obtained with $\lambda = 0.05$, which is a value that prioritises the training of the change detection branch of the network. We then see that the same network trained with sequential training (Str. 4.2) obtained even better results in both change detection and land cover mapping without needing to search for an adequate parameter $\lambda$. This, according to our results, is the best semantic change detection method. By comparing the results for Strategies 3 and 4 we can see the improvements that result directly from integrating the change detection and land cover mapping branches of the networks. In other words, Strategy 4.2 allows us to maximise the change detection performance without reducing the land cover mapping accuracy.

The best performing land cover mapping method was the single purpose network that was trained and used for Strategies 1 and 3. The fact that it achieves a better kappa coefficient than Strategy 4.2 is merely due to the randomness of the initialisation and training of the network, as the land cover mapping branches of Strategy 4.2 are identical to those used in Strategies 1 and 3. This also explains why their results are so similar. By comparing these results to those of Strategy 4.1 it emphasises once again the fact that attempting to train the network shown in Fig.~\ref{fig:frankennet} all at once damages performance in both change detection and land cover mapping.

In Fig.~\ref{fig:results} we can see the results of the proposed networks on a pair of images from the dataset. Note the amount of false detections by Strategy 1 due to the lack of accuracy of prediction of the land cover maps on region boundaries. The second row shows the predicted classes at each pixel for each image. The semantic information about the changes comes from comparing these two predictions. For example, comparing the images in Fig.~\ref{fig:results} (k) and (o) we can say that the changes predicted in (g) were from the "Agricultural areas" class to the "Artificial surfaces" class.

\begin{table*}[t]
    \caption{\label{tab:hrscd-results}Change detection (CD) and land cover mapping (LCM) results of all four of the proposed strategies on the HRSCD dataset. Comparison with the methods proposed by \citet{el2016convolutional} (Otsu [CNNF-O] and fixed [CNNF-F] thresholding) and by \citet{celik2009unsupervised} ([PCA+KM]) are included. Results are in percent.}
    \centering
    \begin{tabular}{|c||c|c|c||c|c|}
        \hline
         & \multicolumn{3}{|c||}{CD} & \multicolumn{2}{c|}{LCM}\\ \hline
         & Kappa & Dice & Tot. acc. & Kappa & Tot. acc. \\ \hline \hline
        Str. 1 & 3.99 & 5.56 & 86.07 & \textbf{71.92} & 87.22 \\ 
        Str. 2 & 21.54 & - & \textbf{98.30} & - & - \\  
        Str. 3 & 12.48 & 13.79 & 94.72 & \textbf{71.92} & 87.22 \\ 
        Str. 4.1 & 19.13 & 20.23 & 96.87 & 67.25 & 85.74 \\
        Str. 4.2 & \textbf{25.49} & \textbf{26.33} & 98.19 & 71.81 & \textbf{89.01} \\   \hline
        \footnotesize{CNNF-O} & 0.74 & 2.43 & 64.54 & - & - \\
        \footnotesize{CNNF-F} & 3.28 & 4.84 & 88.66 & - & - \\
        \footnotesize{PCA+KM} & 0.67 & 2.31 & 83.95 & - & - \\   \hline
    \end{tabular}
\end{table*}

In our tests we observed that the trained networks had the tendency to overestimate the size of the detected changes. It is likely that this happens simply due to the nature of the data that was used for training. The labels in the HRSCD dataset, which come from Urban Atlas, mark as a change the whole terrain where a change of class happened. This means that not only the pixels associated with a given change are marked as change, but the neighbouring pixels that are in the same parcel are also marked as change. This leads to the networks learning to overestimate the boundary of the detected changes in an attempt to also correctly classify the pixels surrounding the detected change. This once again reflects the challenges of the HRSCD dataset.

The performance of two state-of-the-art CD methods are also shown in Table~\ref{tab:hrscd-results}. The first method, proposed by \citet{el2016convolutional}, is based on transfer learning and uses features from a pretrained VGG-19 model~\citep{vgg} to create pixel descriptors, whose Euclidean distance is used to build a difference image. The original method uses Otsu thresholding to perform CD, but we have found that such approach leads to overestimating changes. We therefore tuned a fixed threshold (T = 2300) using a few example images and used that value to test the algorithm on all test data, which significantly increased its performance by reducing false positives. Also included are the results by the method proposed by \citet{celik2009unsupervised}, which performs principal component analysis (PCA) and k-means clustering on the pixels to detect changes in an unsupervised manner. Both algorithms perform worse than the proposed method on the HRSCD dataset.

To evaluate the size of the dataset, we have also tested Strategy~4.2 using reduced amounts of data for training the network. The kappa coefficient, in percent, obtained by using the whole training dataset is $25.49$. This value is reduced to $23.34$ by using half the training data, and is further reduced to $22.18$ by using a quarter of the data. This shows that, as expected, using more data for training the network leads to better results. Nonetheless, it also shows that the dataset is large enough to allow for even more complex and data hungry methods to be trained using the HRSCD dataset in the future.

Finally, it is important to note that the label imperfections in the HRSCD dataset occur not only in the training images, but also in the test images. This means that the performance of the proposed methods may be even higher than the numbers suggest, since some of the disagreements between prediction and ground truth data are actually due to errors in the ground truth data.

\subsection{Eppalock lake images}

\begin{table}[t]
    \caption{\label{tab:overfitting}Change detection results on Eppalock lake test images. Results are in percent.}
    \centering
    \begin{tabular}{|cc|c|c|}
        \hline
         &  & ReCNN-LSTM & EF \\  \hline \hline
       \parbox[t]{2mm}{\multirow{4}{*}{\rotatebox[origin=c]{90}{Binary CD}}}  &  Tot. acc. & 98.67 & \textbf{99.35}  \\   
         & Kappa & 97.28 & \textbf{98.67} \\   
         & No change & 98.83 & \textbf{99.47} \\   
         & Change & 98.46 &  \textbf{99.19} \\   \hline \hline
       \parbox[t]{2mm}{\multirow{6}{*}{\rotatebox[origin=c]{90}{Semantic CD}}}  &  Tot. acc. & \textbf{98.70} &  98.48 \\   
         & Kappa & \textbf{97.52} & 97.10 \\   
         & No change & \textbf{98.49} & 97.73 \\   
         & City exp. & 84.72 & \textbf{100}  \\  
         & Soil change & \textbf{100} &  86.07 \\  
         & Water change & 99.25 & \textbf{99.93}  \\   \hline 
    \end{tabular}
\end{table}

We compare our method in this section to the one proposed by \citet{mou2018learning}, which used recurrent convolutional neural networks for change detection. In that work, pixels were randomly split into train and test sets. We believe that this split leads to overfitting since neighbouring pixels contain redundant information. This is especially true when using CNNs, which take as inputs patches centred on the considered pixels, meaning the network sees the same information for training and testing. It is likely that overfitting takes place, since an accuracy of over 98\% is achieved by using only 1000 labelled pixels to train a network with 67500 parameters (for their  long short-term memory (LSTM) architecture, which performed the best). The data consists of a single image pair of 631x602 pixels only partially annotated, with a total of 8895 annotated pixels which is much less data than what is required for deep learning methods. The HRSCD dataset presented in Section~\ref{sec:dataset} contains over 3 million times more labelled pixels than the Eppalock lake image pair. Despite the flaws of this testing scheme, we have followed it to achieve a fair comparison between the methods.

Using the CNN architecture labelled EF by \citet{daudt2018urban}, we have achieved excellent numeric results which discouraged the usage of more complex methods which would lead to even more extreme overfitting. The results achieved by the EF network were better for binary change detection and equivalent for semantic change detection compared to ReCNN-LSTM. The results can be seen in Table~\ref{tab:overfitting}.

\section{Conclusion}

The first major contribution presented in this paper is the first large scale very high resolution semantic change detection dataset that will be released to the scientific community. This dataset contains 291 pairs of aerial images, together with aligned rasters for change maps and land cover maps. This dataset allows for the first time for deep learning methods to be used in this context in a fully supervised manner with minimal concern for overfitting.
We have then proposed different methods for using deep FCNNs for semantic change detection. The best among the proposed methods is an integrated network that performs land cover mapping and change detection simultaneously, using information from the land cover mapping branches to help with change detection. We also proposed a sequential training scheme for this network that avoids the need of tuning a hyperparameter, which circumvents a costly grid search.

The automatic methods used to generate the HRSCD dataset resulted in noisy labels for both training and testing, and how to deal with this problem is still an open question. It would also be interesting to explore ways to explicitly deal with parallax problems which are present in VHR images which sometimes lead to false positives due to the different points of view and the geometry of the scene.

\section*{Acknowledgments}

This work is part of ONERA's project DELTA. We thank X. Zhu and L. Mou (DLR) for the Eppalock Lake images.



\bibliographystyle{model2-names}
\bibliography{refs}

\end{document}